\documentclass[letterpaper, 10 pt, journal, twoside]{ieeetran}

\IEEEoverridecommandlockouts                              

\usepackage{graphics} 
\usepackage{epsfig} 

\usepackage{amsmath} 
\usepackage{amssymb}  
\usepackage{lipsum}
\usepackage{mathtools}
\usepackage{cuted}
\usepackage{arydshln}
\usepackage{subcaption}
\usepackage{hyperref}

\usepackage{color}
\usepackage{booktabs}
\usepackage{multirow}  
\usepackage{tabularx}
\usepackage{caption}
\usepackage{subcaption}
\usepackage{cite}
\usepackage{ulem}

\title{Contact-Aware Incremental Model Predictive Control for an Underactuated Aerial Manipulator}
\author{Darwin Liu, Tamas Keviczky, Sihao Sun$^*$ \thanks{The authors are with the Faculty of Mechanical Engineering, Delft University of Technology.} \thanks{${}^*$ Corresponding Author: s.sun-2@tudelft.nl}}
\IEEEaftertitletext{\vspace{-2.5\baselineskip}}
\begin{document}
\maketitle

\begin{abstract}
We present a robust contact-aware control framework for aerial writing on an underactuated platform. The framework combines nonlinear model predictive control (NMPC) for accurate end-effector position and normal-force tracking at small reference penetration depths, with consistent performance across controller tunings, with whole-body incremental nonlinear dynamic inversion (INDI) for robustness to frictional and aerodynamic disturbances during contact.
The proposed controllers are validated on a quadrotor-based aerial manipulator with a rigid, single-link, one-degree-of-freedom (DoF) arm in simulation and real-world experiments.
The aerial writing experiments span vertical and inclined surfaces, multiple reference forces, different friction conditions, and wind disturbances. The results demonstrate that robust simultaneous five-DoF end-effector pose and contact-force tracking is achievable on a standard underactuated quadrotor with a simple, rigid, single-link arm, without requiring a fully actuated platform, a complex arm, or dedicated force/torque sensing.
\end{abstract}



\section{Introduction}\label{sec:introduction}
Unmanned Aerial Vehicles (UAVs) are increasingly deployed beyond sensing and inspection tasks, towards active physical interaction with the environment. Aerial manipulators, UAVs equipped with a robotic arm or end-effector, enable a broad range of manipulation tasks, including maintenance \cite{bodie_omnidirectional_2019}, assembly \cite{dong_centimeter-level_2022}, opening doors \cite{cuniato_learning_2023}, harvesting \cite{gupta2025umi}, and surface interaction tasks such as sliding and writing \cite{tzoumanikas_aerial_2020, bodie_active_interaction_2021, zhang_learning_2022, guo2024flying}. 
These tasks require simultaneous control of end-effector pose and contact force, and place demanding requirements on the robustness and accuracy of the underlying control system. Benchmarks have been designed to compare performance across different aerial manipulator morphologies \cite{suarez2020benchmarks}.

Underactuated aerial manipulators typically offer the greatest mechatronic simplicity among various morphologies.
Yet achieving accurate and robust control for an underactuated aerial manipulator is challenging due to the inherent nonlinearity of the system dynamics, unmodeled disturbances such as friction and aerodynamic effects, and the dynamic coupling between the aerial base and the manipulator \cite{ollero_past_2022}. Nonlinear Model Predictive Control (NMPC) has emerged as a leading approach for aerial manipulation, offering the ability to incorporate system constraints and handle complex coupled dynamics \cite{tzoumanikas_aerial_2020, lee_aerial_2020}. However, NMPC alone lacks the reactivity to reject fast unmodeled disturbances, which is a critical limitation in contact tasks involving friction or wind. 

Incremental Nonlinear Dynamic Inversion (INDI) is a sensor-based nonlinear control method that has demonstrated strong disturbance rejection properties in quadrotor flight \cite{smeur_cascaded_2018, sun_comparative_2024}. INDI directly exploits onboard measurements to compensate for model uncertainty and external disturbances, reducing its reliance on an accurate system model. Recent work has shown that combining an outer-loop NMPC with an INDI inner loop significantly improves tracking performance and robustness in agile quadrotor flight \cite{sun_comparative_2024}. Despite its demonstrated effectiveness in quadrotor control, the application of INDI to aerial manipulators is limited \cite{deshmukh2025global, park_adaptive_2025}. These works only account for UAV base dynamics, and lack a whole-body formulation for multi-bodied aerial manipulators. Furthermore, the potential of INDI in contact-rich aerial manipulation tasks has yet to be explored.

\begin{figure}[t!]
    \centering
\includegraphics[width=\linewidth]{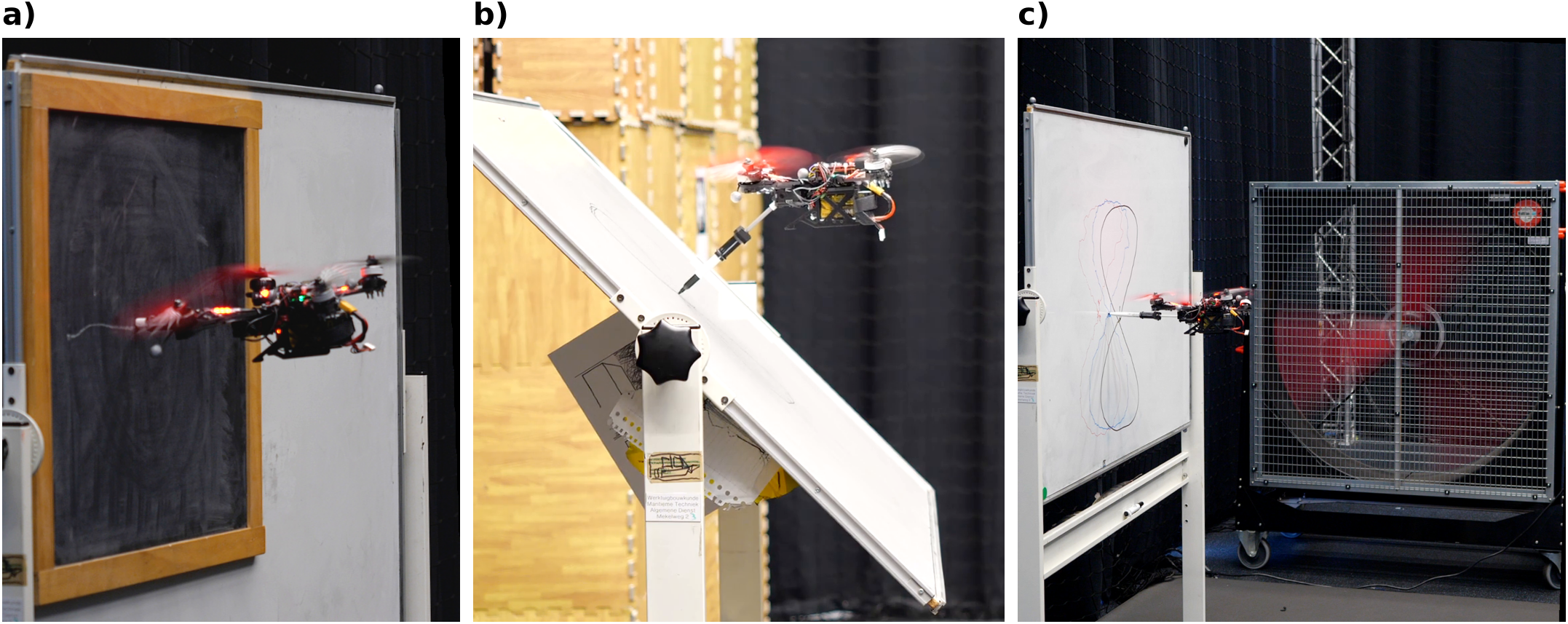}
    \caption{Aerial writing snapshots. a) Chalk on blackboard. b) Marker on inclined whiteboard. c) Marker on vertical whiteboard with external wind.}
    \label{fig:eyecatcher}
\end{figure}

Beyond control challenges, high hardware complexity remains a major constraint in aerial manipulation. Existing contact-sliding frameworks mostly rely on fully omnidirectional bases \cite{bodie_active_interaction_2021, zhang_learning_2022, guo2024flying}, trading structural simplicity for control authority. 
Underactuated alternatives \cite{tzoumanikas_aerial_2020, meng2019hybrid} exist, but are restricted to horizontal interactions and cannot achieve tracking across varying surface orientations.
Furthermore, relying on dedicated force/torque or tactile sensors \cite{nava_direct_2020, guo2024aerial} incurs significant mass, cost, and calibration penalties. It remains an open question whether a standard underactuated quadrotor with a simple, rigid single-link arm can achieve accurate simultaneous pose and force tracking without dedicated force sensing.

In this work, we address both of these gaps. We propose a contact-aware NMPC formulation (CA-NMPC) for an underactuated aerial manipulator.
This is combined with a whole-body INDI inner loop to improve robustness against unmodeled disturbances. 
The proposed controllers are validated on an aerial manipulator composed of a quadrotor-based platform with a one-DoF non-compliant arm, in both simulation and real-world aerial writing experiments on surfaces of different orientations and friction properties, and under external wind disturbances (Fig.~\ref{fig:eyecatcher}).

The main contributions of this paper are threefold:
\begin{itemize}
    \item A contact-aware NMPC formulation for non-compliant underactuated aerial manipulation with no dedicated force/torque sensing.
    \item A cascaded whole-body INDI controller to robustify the proposed NMPC against unmodeled disturbances.
    \item Validation in simulation and real-world experiments of the proposed controllers on aerial writing tasks on vertical and inclined surfaces under different friction and wind disturbances.
\end{itemize}

\section{Preliminaries}\label{sec:prelim}

\subsection{Notation}
Throughout this letter, scalars are denoted by lowercase non-bold letters, vectors by lowercase bold letters, and matrices by uppercase letters. The inertial, quadrotor-body, and arm frames are denoted by $\Sigma_I$, $\Sigma_B$, and $\Sigma_A$, respectively. A vector expressed in $\Sigma_A$ is written as $\prescript{}{A}{\boldsymbol{v}}$; vectors without an explicit frame designation are expressed in $\Sigma_I$. The relative-position vector $\boldsymbol{r}_{ab}$ points from point $a$ to point $b$. The relevant points are the quadrotor-base center of mass $b$, the arm root $r$, and the end-effector $e$.

\subsection{Multibody Kinematics and Dynamics}
The aerial manipulator is modeled as a floating multibody system consisting of two rigid bodies, the quadrotor base and the arm with its end-effector, connected by a one-DoF hinge joint that allows the arm to rotate about the $\boldsymbol{y}_B$ axis. This morphology enables five-DoF end-effector pose tracking; only rotation about the arm axis remains uncontrolled.

\begin{figure}[t]
    \centering
    \includegraphics[width=0.9\linewidth]{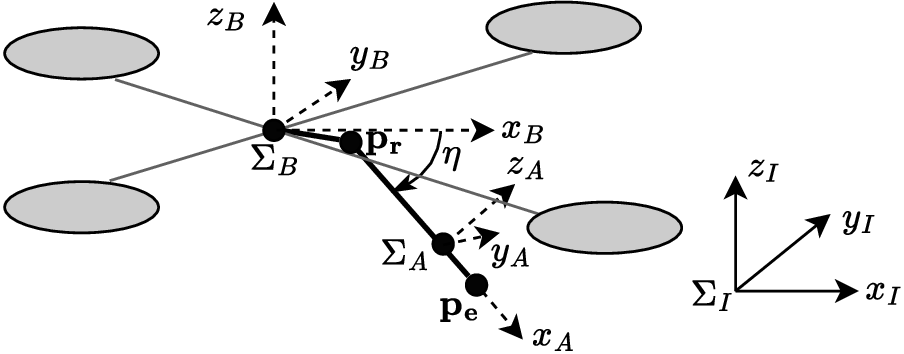}
    \caption{A schematic of the single-link underactuated aerial manipulator with the relevant reference frames and coordinates.}
    \label{fig:schematic_DSAM}
\end{figure}

A schematic of the aerial manipulator is shown in Fig.~\ref{fig:schematic_DSAM}. 
We introduce a set of generalized coordinates for the kinematic and dynamic models.
The coordinates comprise the base position $\boldsymbol{p}_b \in \mathbb{R}^3$, the base-orientation quaternion $\boldsymbol{q}_b = [q_w, q_x, q_y, q_z]^\top \in \mathbb{S}^3$, and the arm joint angle $\eta\in [-\frac{2}{3}\pi,~\frac{2}{3}\pi]$. The generalized-coordinate vector is
\begin{equation}
    \boldsymbol{q} = [\boldsymbol{p}_b^\top, \; \boldsymbol{q}_b^\top, \; \eta]^\top.
\end{equation}

The end-effector position in the inertial frame is
\begin{equation}
    \boldsymbol{p}_e = \boldsymbol{p}_b + R_{IB}(\boldsymbol{q}_b)(\prescript{}{B}{\boldsymbol{r}}_{br}  + R_{BA}(\eta)\prescript{}{A}{\boldsymbol{r}}_{re}),
\end{equation}
where $R_{IB}(\boldsymbol{q}_b)$ is the rotation matrix from $\Sigma_B$ to $\Sigma_I$, and $R_{BA}(\eta)$ is the rotation matrix from $\Sigma_A$ to $\Sigma_B$.

The generalized velocities $\boldsymbol{v}$ comprise the inertial linear velocity of the base $\boldsymbol{v}_b$, the base angular velocity expressed in $\Sigma_B$, $\prescript{}{B}{\boldsymbol{\omega}}_b$, and the joint velocity $\dot{\eta}$. Thus,
\begin{equation}
    \boldsymbol{v} = [\boldsymbol{v}_b^\top  ,~ \prescript{}{B}{\boldsymbol{\omega}}_b^\top  ,~ \dot{\eta}]^\top  \in \mathbb{R}^{7}.
\end{equation}

The generalized-coordinate derivatives satisfy
\begin{equation}
    \dot{\boldsymbol{q}} = E(\boldsymbol{q}) \boldsymbol{v} = \text{blkdiag}\left(\mathbb{I}_{3 \times 3}, \frac{1}{2}\Lambda(\boldsymbol{q}_b), 1\right) \boldsymbol{v},
\end{equation}
where $\Lambda(\boldsymbol{q}_b)\in\mathbb{R}^{4\times 3}$ is the standard quaternion propagation matrix for a body-frame angular velocity \cite{siciliano_springer_2008}.

Constraint-consistent dynamics are derived using the projected Newton--Euler method \cite{siciliano_springer_2008}. The multibody dynamics are
\begin{equation}\label{eq:multibody_dynamics}
    M(\boldsymbol{q})\dot{\boldsymbol{v}} + C(\boldsymbol{v}, \boldsymbol{q})\boldsymbol{v} + \boldsymbol{g}(\boldsymbol{q}) = \boldsymbol{\tau} +
    J_{E}^\top \boldsymbol{f}_{c},
\end{equation}
where $M(\boldsymbol{q}),C(\boldsymbol{v},\boldsymbol{q})\in\mathbb{R}^{7\times7}$ are the mass and Coriolis/centrifugal matrices, respectively, and $\boldsymbol{g}(\boldsymbol{q})\in\mathbb{R}^7$ is the gravity vector. The generalized actuation $\boldsymbol{\tau}=[\boldsymbol{t}^\top,\boldsymbol{\tau}_b^\top,\tau_\eta]^\top$ comprises the inertial-frame thrust $\boldsymbol{t}=R_{IB}(\boldsymbol{q}_b)\prescript{}{B}{\boldsymbol{t}}$, the body-frame torque $\boldsymbol{\tau}_b$, and the arm torque $\tau_\eta$. Here, $J_E\in\mathbb{R}^{3\times7}$ is the translational end-effector Jacobian, and $\boldsymbol{f}_c$ is the inertial-frame contact force.



\subsection{Task-Space Planning}
An aerial writing task is specified by the reference end-effector pointing direction $\boldsymbol{x}_{A_r} = R_{IA_r} [1\; 0\; 0]^\top$, end-effector position $\boldsymbol{p}_{e,r}$, and contact force $\boldsymbol{f}_{c,r}$. Here, $R_{IA_r}$ is the desired rotation from the reference arm frame $\Sigma_{A_r}$ to $\Sigma_I$, and the subscript $r$ denotes a reference quantity. 
A global planner exploits the task-space flatness of the aerial manipulator, under the assumption $\boldsymbol{r}_{br} = 0$, to derive a reference configuration-space trajectory~\cite{welde_dynamically_2021} while accounting for the effect of the desired contact force on the required thrust direction.
Differential flatness does not hold exactly for our morphology. We mitigate this discrepancy by assigning high weights to the end-effector cost and low weights to the configuration-space cost in the NMPC (Sec.~\ref{ssec:exp_setup}). The purpose of the planner is to provide a smooth end-effector pose and force reference, together with the approximate generalized velocities required for tracking.

\section{Methodology}\label{sec:method}


\subsection{Baseline NMPC Formulation}\label{ssec:baseline_mpc}
The baseline NMPC extends the quadrotor agile-flight NMPC of \cite{sun_comparative_2024} to the multibody aerial manipulator dynamics (\ref{eq:multibody_dynamics}), without modeling contact forces, by updating the dynamic constraints and adding cost terms for end-effector position, attitude, and joint tracking.

The state vector $\boldsymbol{x}$ in the MPC consists of the generalized coordinates $\boldsymbol{q}$ and generalized velocities $\boldsymbol{v}$. 
The input vector is $\boldsymbol{u} = [\boldsymbol{\omega}_p^\top  \; {\eta}_\text{cmd}]^\top$, where $\boldsymbol{\omega}_p$ is the vector of propeller speeds and ${\eta}_\text{cmd}$ is the reference joint position in the low-level PID controller of the joint actuators. 
Due to hardware constraints, our system will not use torque control but instead position control for the manipulator. 
The dynamics of the arm in the whole-body dynamics and thus the NMPC, are approximated as:
\begin{equation}
        \ddot{\eta} = k_{p, \text{arm}} (\eta_{\text{cmd}} - \eta) - k_{d,\text{arm}} \dot{\eta}.
\end{equation}
Furthermore, the propeller speeds $\boldsymbol{\omega}_p$ also need to be mapped to body torques and forces for the multibody dynamics. 
For the MPC formulation, a control allocation is used where only the drag and thrust of each propeller is accounted for:
\begin{equation}\label{eq:control_allocation}
        \left[\boldsymbol{\tau}_b^\top,
         \prescript{}{B}{\boldsymbol{t}}^\top \right]^\top =
    H (\boldsymbol{\omega}_p \circ \boldsymbol{\omega}_p)
\end{equation}
with
$\circ$ the Hadamard product and $H \in \mathbb{R}^{6\times4}$ the control allocation matrix described in \cite{bicego2020nonlinear}; $\boldsymbol{\tau}_b$ and $\prescript{}{B}{\boldsymbol{t}}$ are expressed in $\Sigma_B$, with $\boldsymbol{t}=R_{IB}(\boldsymbol{q}_b)\prescript{}{B}{\boldsymbol{t}}$ in \eqref{eq:multibody_dynamics}.
The following errors for stage and terminal costs are defined:
\begin{equation}
\begin{gathered}
    \boldsymbol{e}_{p,B} = \boldsymbol{p}_b - \boldsymbol{p}_{b,r}, \qquad
    \boldsymbol{e}_v = \boldsymbol{v}_b - \boldsymbol{v}_{b,r},  \\
    \boldsymbol{e}_{q,B} = [\boldsymbol{q}_{b}^{-1} \otimes \boldsymbol{q}_{b,r}]_{x:z}, \qquad
    \boldsymbol{e}_{\omega} = \prescript{}{B}{\boldsymbol{\omega}}_b
    - R_{BB_r} \prescript{}{B_r}{\boldsymbol{\omega}}_{b,r},\\
    e_\eta = \eta- \eta_r, \qquad
    e_{\dot{\eta}} = \dot{\eta} - \dot{\eta}_r, \\
    \boldsymbol{e}_{p,E} = \boldsymbol{p}_e - \boldsymbol{p}_{e,r}, \qquad
    \boldsymbol{e}_{n,E} =
    \boldsymbol{x}_A - \boldsymbol{x}_{A_r},\\
    \boldsymbol{u}_\text{eff} = [\boldsymbol{\omega}_p^\top \; \ddot{\eta}]^\top 
\end{gathered}
\end{equation}
Here $\boldsymbol{x}_{A}=R_{IB}(\boldsymbol{q}_b)R_{BA}(\eta)[1\;0\;0]^\top$, $R_{BB_r}=R_{IB}(\boldsymbol{q}_b)^\top R_{IB}(\boldsymbol{q}_{b,r})$, and $[\,\cdot\,]_{x:z}$ extracts the vector part of a quaternion.
Each of the errors has a corresponding weighting matrix $Q_{p,B}$, $Q_{v}$, $Q_{q,B}$, $Q_{\omega}$, $Q_\eta$, $Q_{\dot{\eta}}$, $Q_{p,E}$ and $Q_{n,E}$, which are parameters that have to be chosen. The control-effort vector $\boldsymbol{u}_\text{eff}$ is weighted by $Q_u$.

This optimization problem is solved each time step in a receding horizon fashion. The optimal input sequence obtained each timestep will be the desired rotor speed commands and the desired joint angles of the arm.

The baseline NMPC does not model environmental interaction forces. How this affects force tracking will be explored in Sec.~\ref{sssec:pen_depth}.

\subsection{Contact-Aware NMPC Formulation}
The contact-aware NMPC formulation proposed in this work, hereafter \textit{CA-NMPC}, incorporates contact modeling into the dynamic constraints and cost. An a priori surface model provides a point $\hat{\boldsymbol{p}}_c$ on the surface and an outward unit normal $\hat{\boldsymbol{e}}_n$, directed from the surface into free space. 

For force tracking, the cost function penalizes
\begin{equation}\label{eq:force_tracking_error}
    e_{f,k} = \|\boldsymbol{f}_{c,\text{cost},k}\|_2 - \|\boldsymbol{f}_{c,r,k}\|_2 ,
\end{equation}
the deviation between a predicted contact force $\boldsymbol{f}_{c,\text{cost},k}$ and the desired reference force $\boldsymbol{f}_{c,r,k}$, weighted by $Q_f$. The predicted force is in turn modeled as
\begin{equation}\label{eq:contact_force_cost}
    \boldsymbol{f}_{c,\text{cost},k} = \max\left(0,\; i_{c,k}\, \hat{f}_n + k_{p,\text{c}}\, d_{\text{pen},k} \right) \hat{\boldsymbol{e}}_n ,
\end{equation}
i.e., the current contact normal force estimate $\hat{f}_n$ multiplied by a contact activation variable $i_{c,k}$ described later, corrected by a term proportional to a signed distance between the contact surface and the end-effector $d_{\text{pen},k} = (\hat{\boldsymbol{p}}_c - \boldsymbol{p}_{e,k})^\top \hat{\boldsymbol{e}}_n$ we call the \textit{virtual penetration depth}, scaled by a contact-stiffness parameter $k_{p,\text{c}}$. 
$k_{p,\text{c}}$ is a force-per-length tuning gain and not a real stiffness, set empirically to balance tracking accuracy against the oscillatory contact behavior induced by excessively high values. $d_{\text{pen},k}$ is not a real displacement either: since the end effector is non-compliant and never penetrates the surface, it is purely a modeling device. This contrasts with a compliant end-effector as in e.g.~\cite{tzoumanikas_aerial_2020}, where this virtual penetration depth could correspond to an actual spring compression.

A similar but distinct modeling variable is the \textit{reference} penetration depth $d_{\text{pen},r,k} = (\hat{\boldsymbol{p}}_c - \boldsymbol{p}_{e,r,k})^\top \hat{\boldsymbol{e}}_n$, evaluated against the task-space planner's reference trajectory $\boldsymbol{p}_{e,r,k}$.
As the reference trajectory approaches the surface, the reference force is ramped in smoothly over a contact transition boundary layer of width $w_t$, using a contact activation variable $i_{c,k} \in [0,1]$. The variable $i_{c,k}$ follows a smoothstep-like profile of $d_{\text{pen},r,k}$: it is $0$ for $d_{\text{pen},r,k} \le -w_t$, rises smoothly to $1$ as $d_{\text{pen},r,k}$ increases to $0$, and remains $1$ thereafter, with chosen $w_t = 2$~cm. Note that $i_{c,k}$ is planned from the reference trajectory and prior contact surface information, so activation occurs at a fixed point in the task-space plan.

The contact force entering the MPC dynamics constraint \eqref{eq:multibody_dynamics} is defined separately from the cost-function force as $\boldsymbol{f}_{c,\text{dyn},k} = i_{c,k}\, \boldsymbol{f}_{c,r,k}$, directly using the \textit{reference} force $\boldsymbol{f}_{c,r,k}$ as a feedforward term. This choice, rather than a spring model as in \cite{tzoumanikas_aerial_2020}, is deliberate for non-compliant interaction: the prediction model anticipates the desired steady-state contact force regardless of end-effector position, corresponding more to open-loop control. Closed-loop correction of the actual force error is handled through the cost term \eqref{eq:force_tracking_error}, which uses the live normal-force estimate $\hat{f}_n$.

\subsection{External Force Estimator}\label{sec:force_estimator}
The external force estimator assumes quasi-static arm motion during interaction, so arm dynamics can be neglected. The external force estimate in the inertial frame is then given by:
\begin{equation}
     \hat{\boldsymbol{f}}_\text{ext} = R_{IB}(\boldsymbol{q}_b)
    \left( m_{\text{total}}\prescript{}{B}{\boldsymbol{a}}_f - \prescript{}{B}{\boldsymbol{t}}_{f} \right)
    \label{eq:external_wrench_observer}
\end{equation}
where $m_{\text{total}}$ is the total mass, $\prescript{}{B}{\boldsymbol{a}}_f$ the filtered body-frame accelerometer specific force, and $\prescript{}{B}{\boldsymbol{t}}_f$ the body-frame thrust estimated from $\boldsymbol{\omega}_{p,f}$. The normal and in-plane magnitudes are:
\begin{equation}
    \hat{f}_n=\hat{\boldsymbol{f}}_\text{ext}^\top\hat{\boldsymbol{e}}_n,~~~\hat{f}_p=\|\left(\mathbb{I}-\hat{\boldsymbol{e}}_n\hat{\boldsymbol{e}}_n^\top\right)\hat{\boldsymbol{f}}_\text{ext}\|_2
\end{equation}

\begin{figure*}[t]
    \centering
    \includegraphics[width=0.8\linewidth]{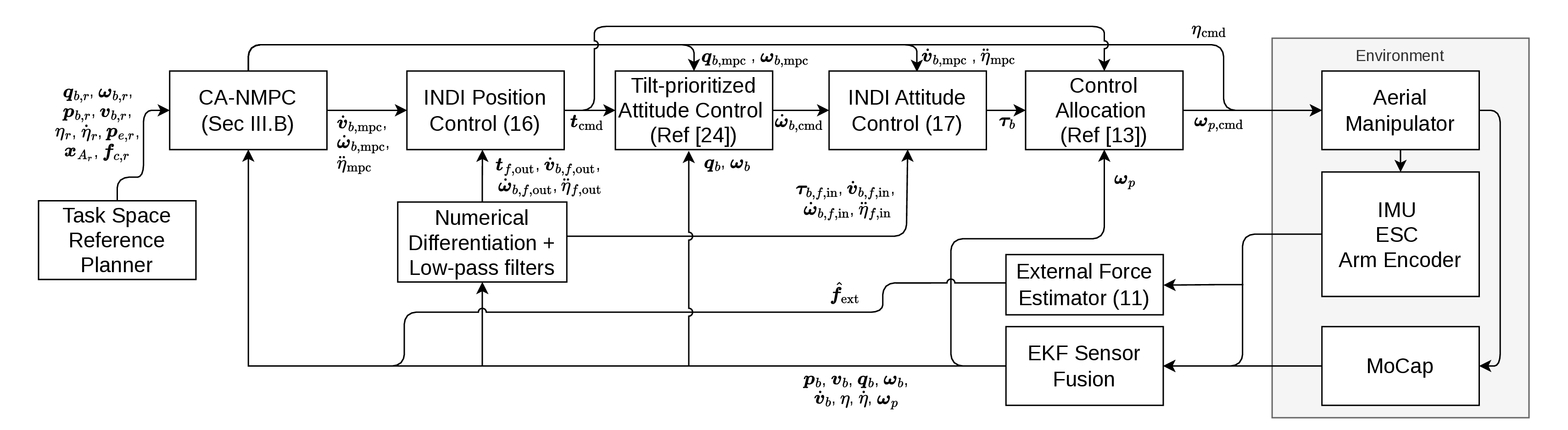}
    \caption{Schematic overview of the proposed CA-NMPC-CINDI controller.
    }
    \label{fig:schematic_nmpc_ff_cindi}
\end{figure*}

\subsection{Cascaded Whole-Body INDI Controller}\label{ssec:whole_body_indi}
The NMPC is robustified with an INDI inner loop that tracks the optimal NMPC trajectory.
Writing the aerial manipulator dynamics~\eqref{eq:multibody_dynamics} as $\dot{\boldsymbol{v}}=\boldsymbol{\phi}(\boldsymbol{x},\boldsymbol{\tau},\boldsymbol{f}_c)$, a first-order Taylor expansion about the previous sample $(\boldsymbol{x}_0,\boldsymbol{\tau}_0,\boldsymbol{f}_{c,0})$ gives
\begin{equation}
    \dot{\boldsymbol{v}} = \dot{\boldsymbol{v}}_0 + M(\boldsymbol{q}_0)^{-1}\left(\boldsymbol{\tau}-\boldsymbol{\tau}_0\right) +
    \boldsymbol{\gamma}(\Delta \boldsymbol{x}, \Delta \boldsymbol{f}_c) + \mathcal{O}(\|\Delta\|^2)
\end{equation}
with $\boldsymbol{\gamma} = \left.\frac{\partial\boldsymbol{\phi}}{\partial\boldsymbol{x}}\right|_0\Delta\boldsymbol{x} + \left.\frac{\partial\boldsymbol{\phi}}{\partial\boldsymbol{f}_c}\right|_0\Delta\boldsymbol{f}_c$ and $\Delta$ collecting all increments. The INDI time-scale-separation assumption, together with slowly varying contact forces, renders $\boldsymbol{\gamma}$ negligible; the latter fails at the instant of contact, as discussed below. Neglecting also the higher-order remainders, replacing $\dot{\boldsymbol{v}}$ with $\dot{\boldsymbol{v}}_\text{mpc}$, obtained by numerical differentiation of the optimal state trajectory at the first predicted timestep, and filtering the previous sampled variables for noise suppression, we get the INDI control law
\begin{equation}\label{eq:whole_body_INDI}
\left[\begin{array}{c}
         \boldsymbol{t}  \\
         \boldsymbol{\tau}_{b}\\
         {\tau}_{{\eta}}
    \end{array}\right]_{\text{cmd}}
    =
\left[\begin{array}{c}
         \boldsymbol{t}_{f}  \\
         \boldsymbol{\tau}_{b,f}\\
         {\tau}_{\eta,f}
    \end{array}\right]
+ 
M(\boldsymbol{q}_f)
\left[\begin{array}{c}
         \dot{\boldsymbol{v}}_{b,\mathrm{mpc}} - \dot{\boldsymbol{v}}_{b,f}  \\
         \prescript{}{B}{\dot{\boldsymbol{\omega}}}_{b,\mathrm{mpc}} - \prescript{}{B}{\dot{\boldsymbol{\omega}}}_{b,f}\\
         \ddot{\eta}_\mathrm{mpc} - \ddot{\eta}_f
    \end{array}\right]
\end{equation}
Here, the subscript $f$ denotes a filtered value. For the derivation of the separate INDI layers, we partition $M \triangleq M(\boldsymbol{q}_f)$ conformally with the block structure above,
\begin{equation}
    M = \begin{bmatrix}
M_{11} & M_{12} & M_{13} \\
M_{12}^\top & M_{22} & M_{23} \\
M_{13}^\top & M_{23}^\top & M_{33}
\end{bmatrix}
\end{equation}
The rigid-body inertial parameters (mass, inertia tensor, and center-of-mass offsets) entering $M$ are measured directly from the physical hardware. Note that $M$ depends on the arm configuration.
The INDI control law requires no model of the external force, Coriolis, or gravitational terms in~\eqref{eq:multibody_dynamics}, and is robust to mismatch in $M$, so that only a coarse inertia estimate is needed~\cite{wang_stability_2019}.

Its validity rests on the time-scale separation underlying INDI: a sampling interval sufficiently small relative to the rate of change of the true dynamics yields asymptotic closed-loop stability~\cite{wang_stability_2019}, bounded in practice by the achievable control frequency and by sensor noise on the accelerations entering the incremental terms. At contact onset this is violated, as $\boldsymbol{f}_c$ rises to a finite value within a time comparable to the sampling interval. This is the same class of perturbation as the sudden actuator faults and structural damage considered in~\cite{wang2019incremental}, where incremental control laws are shown to retain ultimate boundedness under abrupt plant changes.

Directly solving \eqref{eq:whole_body_INDI} would generate physically inadmissible thrust components along the quadrotor’s underactuated directions. 
Thus, a cascaded structure is adopted by splitting \eqref{eq:whole_body_INDI} into two control layers.
The position INDI converts the NMPC desired translational acceleration into a commanded thrust vector $\boldsymbol{t}_\text{cmd}$, from which a tilt-prioritized attitude controller derives a desired angular acceleration for the attitude INDI inner loop. This cascaded scheme is denoted by CA-NMPC-CINDI and summarized in Fig.~\ref{fig:schematic_nmpc_ff_cindi}.

\subsubsection{Position INDI}\label{ssec:position_indi}
The position INDI control law is obtained from the first row of \eqref{eq:whole_body_INDI}:
\begin{equation}\label{eq:position_INDI}
    \begin{aligned}
        \boldsymbol{t}_\text{cmd} = \, &\boldsymbol{t}_{f,\text{out}} + M_{11} \left(\dot{\boldsymbol{v}}_{b,\text{mpc}} - \dot{\boldsymbol{v}}_{b,f,\text{out}}\right) \\
        &+ M_{12} \left(\prescript{}{B}{\dot{\boldsymbol{\omega}}}_{b,\text{mpc}} - \prescript{}{B}{\dot{\boldsymbol{\omega}}}_{b,f,\text{out}}\right) \\
        &+ M_{13} \left(\ddot{\eta}_\text{mpc} - \ddot{\eta}_{f,\text{out}}\right)
    \end{aligned}
\end{equation}
All filtered quantities are evaluated 
using a second-order low-pass filter with a cutoff frequency $f_{\text{cut},\text{out}} =3$~Hz. Equation~\eqref{eq:control_allocation} gives $\prescript{}{B}{\boldsymbol{t}}_{f,\text{out}}$ from the filtered propeller speeds $\boldsymbol{\omega}_{p,f,\text{out}}$.


Given the desired thrust vector $\boldsymbol{t}_\text{cmd}$, a tilt-prioritized attitude controller~\cite{brescianini2018tilt} is used to generate the desired angular acceleration command $\prescript{}{B}{\dot{\boldsymbol{\omega}}}_{b,\text{cmd}}$ for the attitude INDI inner loop (note the small notation difference with the MPC angular acceleration command $\prescript{}{B}{\dot{\boldsymbol{\omega}}}_{b,\text{mpc}}$).

\subsubsection{Attitude INDI}\label{ssec:attitude_INDI}

The attitude INDI layer forms the inner loop of the cascaded architecture. It takes the desired angular acceleration as a virtual input and computes the corresponding desired body torque $\boldsymbol{\tau}_b$.
From the second row of \eqref{eq:whole_body_INDI}, we obtain:

\begin{equation}\label{eq:attitude_INDI}
\begin{aligned}
    \boldsymbol{\tau}_{b,\text{cmd}} 
= &    
        \boldsymbol{\tau}_{b,f, \text{in}}
    +M_{12}^\top
(\dot{\boldsymbol{v}}_{b,\text{mpc}} - \dot{\boldsymbol{v}}_{b,f, \text{in}}) \\
& + M_{22}(\prescript{}{B}{\dot{\boldsymbol{\omega}}}_{b,\text{cmd}} - \prescript{}{B}{\dot{\boldsymbol{\omega}}}_{b,f, \text{in}})
+ M_{23}(\ddot{\eta}_{\text{mpc}} - \ddot{\eta}_{f, \text{in}})
\end{aligned}.
\end{equation}
Note that (\ref{eq:whole_body_INDI}, \ref{eq:position_INDI} and \ref{eq:attitude_INDI}) differ from the INDI control law for a standard quadrotor~\cite{smeur_adaptive_2016, sun_comparative_2024, tal_accurate_2021} as it takes into account the inertial torque due to the quadrotor linear accelerations and arm joint accelerations.
The filtered body torque $\boldsymbol{\tau}_{b,f,\text{in}}$ is again computed from the filtered propeller speeds via \eqref{eq:control_allocation}. All measurements for this layer are filtered using a second-order low-pass filter with a cutoff frequency $f_{\text{cut},\text{in}} =6$~Hz.

Once $\boldsymbol{t}_\text{cmd}$,~$\boldsymbol{\tau}_{b,\text{cmd}}$ are obtained, they are converted to rotor speed commands following the same procedure as a standard quadrotor~\cite{sun_comparative_2024}.
It is worth noting that, based on (\ref{eq:whole_body_INDI}), the joint torque command can also be derived using INDI. However, due to hardware limitations, direct command of the joint torque $\tau_\eta$ to the actuators is not currently feasible, and is therefore left for future work. The desired joint position $\eta_\text{cmd}$ is instead directly taken from the MPC.
\section{Experiments}\label{sec:experiments}
The experimental evaluation is designed to answer the following research questions:
\begin{itemize}
    \item How does the proposed method compare in simulation with the state-of-the-art contact-aware controllers proposed in~\cite{meng2019hybrid, tzoumanikas_aerial_2020} for similar underactuated aerial manipulators?
    \item How does the contact modeling in the NMPC influence the normal force tracking error and its sensitivity to different reference penetration depths?
    \item Is the proposed control design able to perform simultaneous five-DoF end-effector pose and force tracking on surfaces of different orientations?
    \item What is the effect of each INDI control layer on end-effector position and force tracking performance in the presence of unmodeled external disturbances, specifically unknown friction forces during sliding and lateral wind disturbances?
\end{itemize}
The last question is addressed by introducing an additional controller variant in which the desired thrust is taken directly from the NMPC prediction, bypassing the outer INDI position loop.
Because this variant retains only the attitude-INDI layer, we denote it by CA-NMPC-ATT-INDI. We compare it with the full CA-NMPC-CINDI controller (ours) to isolate the contribution of the outer position-INDI layer to disturbance rejection.

\subsection{Simulation and Experimental Setup}\label{ssec:exp_setup}

We evaluated the controllers in simulation and real-world experiments.
The simulations were performed in Simscape (MATLAB)~\cite{mathworks2021simscape}. Normal contact is modeled by a spring-damper law with a smooth transition into contact, and friction is represented by a smoothed Coulomb model with a continuous transition from static to kinetic friction near zero relative velocity. The simulated wall is effectively rigid: its stiffness is six orders of magnitude greater than the virtual contact-stiffness parameter $k_{p,\text{c}}$ used in the CA-NMPC cost function.

Real-world experiments were conducted in an indoor arena equipped with a Vicon motion-capture system. The aerial manipulator uses a flight stack based on Agilicious~\cite{foehn2022agilicious}, running entirely onboard a Raspberry Pi 5 compute module.
The NMPC is implemented using the acados toolbox~\cite{verschueren2022acados} in both simulation and real-world experiments. The NMPC runs at $100$~Hz, and the INDI layers run at $500$~Hz. The NMPC prediction and control horizons are set to $N=20$ with a discretization time of $0.05$~s.

The quadrotor base has a mass of $0.808$~kg. The arm link, including the end-effector pen, has a mass of $0.02$~kg and a length of $20.5$~cm.
The quadrotor has tilted rotors to provide additional yaw authority for sliding tasks. 

The position and orientation of the contact surface are assumed to be known a priori and are provided to the controller as fixed parameters. The CA-NMPC weights and other controller parameters used throughout the experiments are listed in Table~\ref{tab:mpc_params}. The MPC cost is dominated by the weights on the end-effector position, orientation and the yaw angle. The weights on the configuration space, namely the quadrotor base position, the pitch and roll angles, and the joint angle, are negligible in this configuration, but are retained to demonstrate that the controller can also accommodate tasks requiring explicit tracking of the base position or attitude. 
However, we retain a high heading cost to prevent singularities when the arm points vertically downward or upward.

\begin{table}[h]
\centering
\caption{MPC Weights, MPC Parameters and Tilt-prioritized Control Gains \cite{brescianini2018tilt}}
\label{tab:mpc_params}
\begin{tabular}{lc}
\toprule
\textbf{Parameter} & \textbf{Value} \\ 
\midrule
$Q_{p,E}$, $Q_{n,E}$ &  $200\times \mathbb{I}_{3\times3}$,  $200 \times\mathbb{I}_{3\times3}$\\
$Q_{p,B}$    & $\text{diag}([0.2,\ 0.2,\ 0.5])$  \\
$Q_{q,B}$    & $\text{diag}([0.5,\ 0.5,\ 100])$  \\
$Q_v$, $Q_\omega$  & $\mathbb{I}_{3\times3}$, $\mathbb{I}_{3\times3}$  \\
$Q_\eta$, $Q_{\dot{\eta}}$, $Q_{f}$ & $10^{-3}$, $0.1$, $2.0$ \\
$Q_u$                   & $\text{diag}([0.6,\ 0.6,\ 0.6,\ 0.6,\ 1])$  \\
\midrule
$k_{p,\text{arm}}$, $k_{d,\text{arm}}$ & $400$, $30$ \\
$k_{p,\text{c}}$ & $2.0$\\
\midrule
$k_{q,\text{red}}$, $k_{q,\text{yaw}}$ \cite{brescianini2018tilt} & $120$, $80$ \\
$K_\Omega$ \cite{brescianini2018tilt} & $\text{diag}([25, 25, 20])$ \\

\bottomrule
\end{tabular}
\end{table}
\subsection{Simulation Results}
\subsubsection{Reference Penetration Depth Sensitivity}\label{sssec:pen_depth}
To demonstrate the benefit of the proposed MPC contact modeling, this section evaluates steady-state force tracking performance in simulation across varying reference penetration depths. The ground-truth contact forces $f_n$ (normal force) and $f_p$ (friction), are directly available.
We compare our proposed formulation (CA-NMPC-CINDI) with three baselines: the contact-unaware formulation from Sec.~\ref{ssec:baseline_mpc} (NMPC-CINDI), a reimplemented contact-aware MPC~\cite{tzoumanikas_aerial_2020} (CA-NMPC-SOTA) and a reimplemented hybrid impedance controller~\cite{meng2019hybrid} (HIC). Notably, our hardware setup resembles the platform in \cite{meng2019hybrid}, but differs from \cite{tzoumanikas_aerial_2020}, which uses a delta arm and a compliant end-effector.

\begin{figure}[t]
    \centering
    \includegraphics[width=0.9\linewidth]{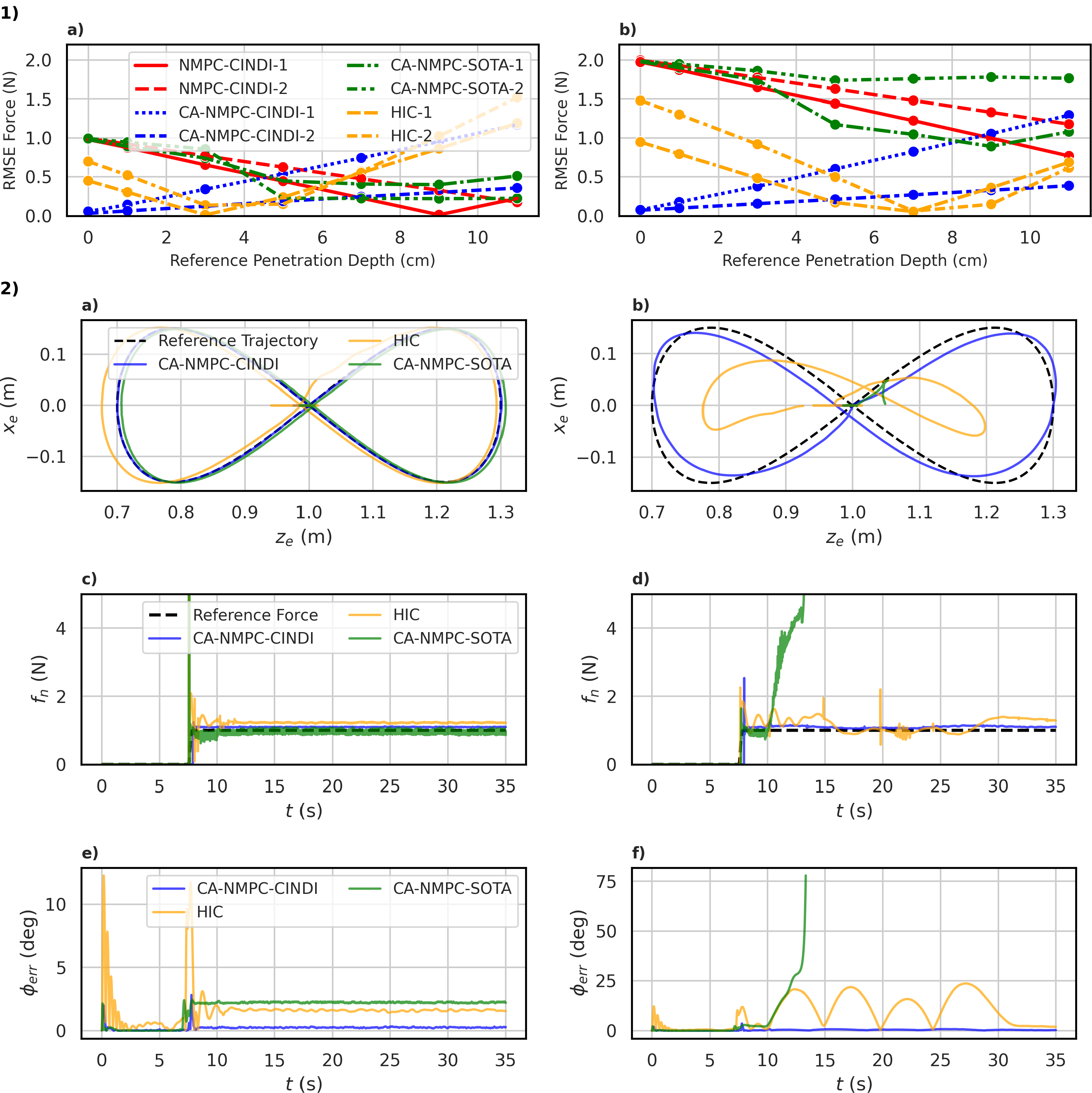}
    \caption{Simulation results. 1) Force tracking RMSE vs. reference penetration depth (point contact, vertical surface) for a) $1$~N reference force, b) $2$~N reference force. 2) Figure-eight writing experiment where subfigures a, c, e are simulated without friction and b, d, f are simulated with friction. a,b) End-effector trajectory projected on the $xz$-plane. c,d) Contact normal force tracking. e,f) Attitude tracking error. }
    \label{fig:sim_results}
\end{figure}

Fig.~\ref{fig:sim_results} compares the force tracking performance of the four different control designs in a point contact experiment on a rigid vertical surface.
Each is evaluated with two sets of weights; the -1 MPC variants share the weights listed in Table \ref{tab:mpc_params}. Since NMPC-CINDI does not track a contact force, it omits $Q_f$. The -2 MPC variants are identical, with the end-effector weights $Q_{p,E}$ and $Q_{n,E}$ changed to $20 \times \mathbb{I}_{3\times3}$.
The reference penetration depth is varied, and the resulting force tracking root mean square error (RMSE) is reported for reference forces of $1$~N and $2$~N. The RMSE is computed over a $5$~s window following the initial contact transient.
The results demonstrate that both CA-NMPC-CINDI controllers achieve accurate force tracking at small reference penetration depths, consistently across different controller tunings and reference forces. 
The baseline controllers, by contrast, exhibit a tuning-dependent optimal penetration depth at which force tracking error is minimized. This sensitivity to tuning is evident from the two distinct curves of each baseline in Fig.~\ref{fig:sim_results}, where the optimal penetration depth differs between two different weight sets and different reference forces. These results highlight a key limitation of the baseline formulations: accurate force tracking requires a reference penetration depth that is matched to specific controller gains and parameters, making it impractical to tune reliably. The proposed contact-aware formulation eliminates this dependency by explicitly incorporating the reference contact force into the dynamics, enabling accurate force tracking at small reference penetration depths regardless of tuning.

\subsubsection{Tracking Accuracy Comparison}\label{sssec:baseline_comp}
This section compares our proposed controller (CA-NMPC-CINDI) with weights in Table \ref{tab:mpc_params} against CA-NMPC-SOTA and HIC, both tuned via Bayesian Optimization~\cite{snoek2012practical} until sufficient performance is reached in a hover and in a static force-tracking experiment. The evaluation function is a weighted sum of end-effector pose and force tracking error and force tracking overshoot. Fig.~\ref{fig:sim_results} shows a task consisting of tracing a figure-eight on a vertical surface with a $f_{n,r} = 1$~N, with and without simulated friction. Without friction, all three controllers achieve accurate force and pose tracking. With friction, however, our proposed method achieves significantly better results. CA-NMPC-SOTA crashes during sliding, after accumulating a large attitude and force tracking error. CA-NMPC-SOTA is not robustified with an INDI inner loop and we observe that NMPC alone is unable to handle significant friction forces. HIC is able to track forces relatively accurately, but performs significantly worse tracking the end-effector position. Due to large attitude deviations during sliding, these two controllers were not implemented in the real-world experiments presented in the following sections.


\begin{table*}[t]
\centering
\caption{Force, in-plane EE (end-effector) position, and EE attitude tracking RMSE and standard deviation, relative to the task-space planner reference, for different trajectories and desired contact forces $f_{n,r}$ during a single experiment. Best value per row in bold. Force tracking RMSE is computed only over time steps where the contact activation exceeds $0.5$. C1--C3: CA-NMPC, CA-NMPC-ATT-INDI, CA-NMPC-CINDI. V.8/I.8: figure-eight drawing on the vertical / $45^{\circ}$-inclined whiteboard.}
\label{tab:combined_rmse_comparison}
\setlength{\tabcolsep}{4pt}
\renewcommand{\arraystretch}{1.1}
\begin{tabular}{llccccccccc}
\toprule
\multirow{2}{*}{Traj.} & \multirow{2}{*}{$f_{n,r}$}
& \multicolumn{3}{c}{Force RMSE (N)}
& \multicolumn{3}{c}{In-plane EE Pos. RMSE (cm)}
& \multicolumn{3}{c}{EE Attitude RMSE (deg)} \\
\cmidrule(lr){3-5} \cmidrule(lr){6-8} \cmidrule(lr){9-11}
& & C1 & C2 & C3 (ours)
  & C1 & C2 & C3 (ours)
  & C1 & C2 & C3 (ours) \\
\midrule
V. 8 & $0.2$
  & $0.32 \pm 0.26$     & $0.072 \pm 0.07$  & $\mathbf{0.063 \pm 0.05}$
  & $4.44 \pm 2.32$    & $\mathbf{1.18 \pm 0.38}$ & $1.96 \pm 0.40$
  & $2.95 \pm 1.98$   & $0.63 \pm 0.23$           & $\mathbf{0.49 \pm 0.25}$ \\
V. 8 & $0.5$
  & $1.04 \pm 0.72$   & $0.07 \pm 0.07$           & $\mathbf{0.06 \pm 0.06}$
  & $33.3 \pm 26.4$   & $2.38 \pm 0.72$           & $\mathbf{1.84 \pm 0.30}$
  & $27.55 \pm 22.03$   & $0.61 \pm 0.30$           & $\mathbf{0.53 \pm 0.22}$ \\
V. 8 & $1$
  & crash                  & $0.07 \pm 0.07$           & $\mathbf{0.06 \pm 0.06}$
  & crash                  & $2.51 \pm 1.13$           & $\mathbf{2.01 \pm 0.51}$
  & crash                  & $\mathbf{0.62 \pm 0.29}$  & $0.71 \pm 0.30$ \\
\addlinespace
I. 8 & $0.2$ & -- & $\mathbf{0.09 \pm 0.07}$ & $\mathbf{0.09 \pm 0.09}$ & -- & $2.03 \pm 0.68$ & $\mathbf{1.37 \pm 0.51}$ & -- & $0.65 \pm 0.33$ & $\mathbf{0.58 \pm 0.27}$ \\
I. 8 & $0.5$ & -- & $0.15 \pm 0.09$ & $\mathbf{0.11 \pm 0.10}$ & -- & $2.29 \pm 0.84$ & $\mathbf{1.52 \pm 0.60}$ & -- & $\mathbf{0.70 \pm 0.38}$ & $0.79 \pm 0.35$ \\
I. 8 & $1$   & -- & $0.16 \pm 0.14$ & $\mathbf{0.14 \pm 0.14}$ & -- & $2.92 \pm 1.01$ & $\mathbf{1.97 \pm 0.87}$ & -- & $0.83 \pm 0.41$ & $\mathbf{0.78 \pm 0.37}$ \\
\bottomrule
\end{tabular}
\end{table*}

\subsection{Real World Experiments: Aerial Writing}\label{ssec:aerial_writing}
We validate our proposed method through real-world experiments, demonstrating the benefits of cascading INDI with contact-aware NMPC under model uncertainties.

\begin{figure}
    \centering
    \includegraphics[width=0.9 \linewidth]{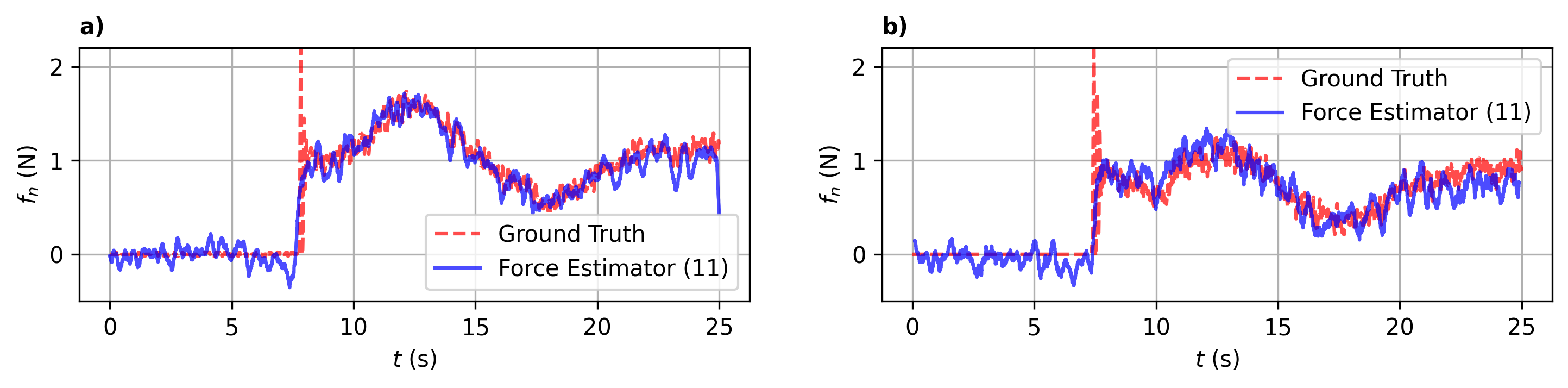}
    \caption{Force tracking validation of force estimator \eqref{eq:external_wrench_observer} against a force/torque sensor attached on a rigid a) vertical and b) $45^\circ$-inclined surface.
    } 
    \label{fig:force_validation}
\end{figure}

With no onboard force sensor, the real-world experiments use $\hat{f}_n$ and $\hat{f}_p$ from \eqref{eq:external_wrench_observer}; plot labels omit the hats. We validate $\hat{f}_n$ on vertical and inclined surfaces using an F/T sensor (Bota SenseOne) attached to the environment, as shown in Fig.~\ref{fig:force_validation}. For the inclined test, estimated rotor-downwash pressure is subtracted from the ground truth.

\subsubsection{Vertical Surface}

This section presents real world aerial writing experiments in which a marker is used to trace trajectories on a vertical whiteboard. 

\begin{figure}
    \centering
    \includegraphics[width=\linewidth]{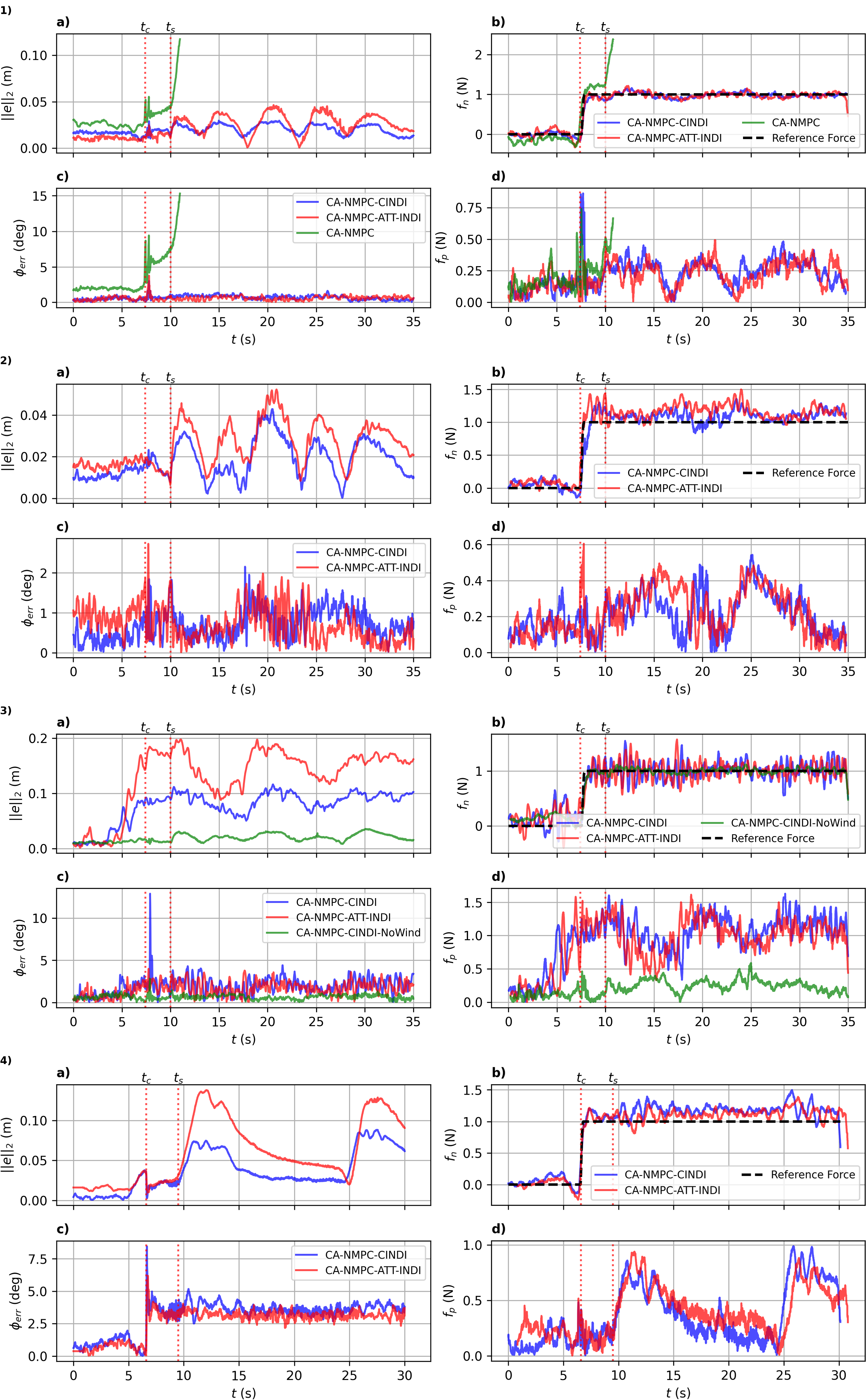}
    \caption{Real world figure-eight writing with $1$~N reference force using a marker on a: 1) vertical whiteboard, 2) inclined whiteboard, 
    and 3) vertical whiteboard with $5$~m/s lateral wind. 4) Sliding on a vertically orientated blackboard with $1$~N reference force using chalk. $t_c$: contact moment, $t_s$: sliding onset. a) End-effector position error. b) Contact force tracking. c) End-effector attitude tracking error. d) Estimated in-plane external-force magnitude $f_p$.}
    \label{fig:experiments}
\end{figure}

The first experiment draws a figure-eight on a vertically oriented whiteboard with a marker with $f_{n,r} = 1$~N, comparing CA-NMPC, CA-NMPC-ATT-INDI, and CA-NMPC-CINDI (Fig. \ref{fig:experiments}). Shortly after the sliding starts, the aerial manipulator with the pure CA-NMPC controller crashed. The INDI controllers are able to complete this task and achieve comparable tracking performance. CA-NMPC-CINDI exhibits slightly more consistent tracking error than CA-NMPC-ATT-INDI in Fig.~\ref{fig:experiments}. Both INDI augmented controllers are able to track the desired force consistently. 

The crash with the CA-NMPC controller can be attributed to friction forces at the end-effector, which induce significant torques on the quadrotor base that the NMPC alone is unable to adequately reject. This result is in-line with our observation in Sec.~\ref{sssec:baseline_comp}. The addition of the INDI attitude inner loop in CA-NMPC-ATT-INDI and CA-NMPC-CINDI provides the necessary disturbance rejection to maintain stable operation in this task, highlighting the importance of the INDI attitude control layer for tasks involving lateral friction forces. 

\subsubsection{Inclined Surface}

Thanks to the one-DoF joint, the aerial writing experiment can also be performed in arbitrary directions. Here we show the experiment of writing the figure-eight on the same whiteboard inclined at $45^{\circ}$ with $f_{n,r} = 1$~N. A snapshot is shown in Fig.~\ref{fig:eyecatcher}. Fig.~\ref{fig:experiments} shows the end-effector pose and force tracking during the experiment of the CA-NMPC-ATT-INDI and CA-NMPC-CINDI controllers. CA-NMPC is excluded from the inclined whiteboard writing tasks because it was not able to produce reliable and consistent writing, owing to its sensitivity to aerodynamic downwash on the surface.

A broader set of aerial writing experiments is summarized in Table~\ref{tab:combined_rmse_comparison}, which report the force, end-effector in-plane position, and end-effector attitude tracking RMSE and standard deviation, across a range of different conditions. These include figure-eight trajectories at different reference contact forces on a vertically oriented whiteboard and the $45^{\circ}$ inclined whiteboard.

Overall, CA-NMPC-ATT-INDI and CA-NMPC-CINDI achieve similar attitude and force tracking performance across experiments. The most notable difference lies in the end-effector position tracking error along the contact surface: CA-NMPC-CINDI exhibits lower tracking error and smaller fluctuations in tracking error during the figure-eight writing experiments. 

\subsection{Robustness During Aerial Writing}

\subsubsection{Wind Disturbance Test}


To evaluate the robustness of the proposed controllers, the figure-eight writing experiment on the vertical whiteboard with $f_{n,r} =1$~N is repeated under lateral wind disturbances of approximately $5$~m/s. A snapshot of the experiment is shown in Fig.~\ref{fig:eyecatcher}.

Fig.~\ref{fig:experiments} shows the end-effector pose and force tracking during the experiment. Note that wind effects are also captured in the estimates of $f_n$ and $f_p$ according to \eqref{eq:external_wrench_observer}, particularly $f_p$, because the wind disturbance is primarily tangential to the contact surface.
The estimated contact normal force remains consistently tracked under wind disturbance, albeit with increased noise. 
As expected, the absolute end-effector tracking errors are significantly larger than in the undisturbed case. Under these conditions, the benefit of the outer INDI position loop becomes clearly apparent: CA-NMPC-CINDI achieves a peak end-effector tracking error of $11$~cm, compared to $20$~cm for CA-NMPC-ATT-INDI, representing a reduction of approximately $45\%$.



\subsubsection{Aerial Writing on a Blackboard}

Aerial writing is next performed using a piece of chalk attached to the end-effector on a blackboard. To the best of our knowledge, this is the first experiment using an aerial manipulator writing on a blackboard using a piece of chalk. This task proves significantly more challenging than whiteboard writing, due to larger and less smooth friction forces arising from transitions between static and kinetic friction regimes. 

The experiment is conducted with $f_{n,r}=1$~N. In Fig.~\ref{fig:eyecatcher} a snapshot of the experiment is shown. Fig.~\ref{fig:experiments} shows the absolute end-effector pose and force tracking for a writing experiment on the blackboard. 

The results demonstrate a clear benefit of the outer INDI position loop in this more demanding contact scenario. CA-NMPC-CINDI achieves a peak end-effector tracking error of $8$~cm during high-acceleration segments, compared to $13$~cm for CA-NMPC-ATT-INDI. In addition, the static tracking offset caused by friction is notably reduced with CA-NMPC-CINDI. Both controllers maintain accurate contact force tracking throughout the experiment. As visible in Fig.~\ref{fig:experiments}, the estimated friction force $f_p$ reaches up to $1$~N, equal in magnitude to $f_{n}$, highlighting the significant frictional disturbances present in this task.


\section{Discussion and Conclusions }\label{sec:conclusions}

This work presents an aerial manipulator that combines a novel contact-aware NMPC with whole-body INDI for simultaneous five-DoF end-effector pose and contact-force tracking. The underactuated platform uses a simple, one-DoF, non-compliant arm. Simulation and real-world aerial writing experiments on vertical and inclined whiteboards, a blackboard, and under an approximately $5$~m/s lateral wind disturbance demonstrate robust force and five-DoF end-effector pose tracking without a compliant end-effector or omnidirectional base.

The results show that the contact-aware extension enables accurate force tracking at small reference penetration depths and remains robust to controller tuning as a clear benefit compared to existing predictive and impedance controllers on a similarly underactuated platform. 
The INDI augmentation proved essential for stable sliding contact. CA-NMPC alone became unstable during sliding because end-effector friction induces significant torques on the quadrotor base that the NMPC was not able to reject. Adding the INDI outer position loop in CA-NMPC-CINDI improved end-effector position tracking across challenging conditions, including lateral wind and high-friction interaction.

While our ground-truth-validated force estimator proved sufficient for our experiments, its sensitivity to unmodeled disturbances motivates future exploration of dedicated sensing or advanced estimation techniques. Additionally, testing our whole-body INDI formulation on a lightweight arm (approximately $3\%$ of total mass) without direct joint torque control left its advantages over base-only variants unconfirmed; investigating heavier manipulators and direct torque actuation remains a key future direction. Finally, our cascaded architecture currently lacks explicit inner-loop actuator constraint handling and exhibits a subtle layer-mismatch issue, where the INDI position layer generates the desired attitude while the disturbance-unaware NMPC provides the angular velocity reference, resulting in an asymptotic tracking offset under sustained disturbances. Incorporating prioritized constraint handling and feeding disturbance estimates directly into the NMPC will be pursued in a future study to reconcile these control layers.

\bibliographystyle{ieeetr}
\bibliography{IEEEabrv,reference}

\end{document}